# LifeCLEF Plant Identification Task 2015

Herv´e Goëau[1], Pierre Bonnet[3], and Alexis Joly[1,2]

[1] Inria ZENITH team, France, name.surname@inria.fr
[2] LIRMM, Montpellier, France
[3] CIRAD, UMR AMAP, France, pierre.bonnet@cirad.fr

**Abstract.** The LifeCLEF plant identification challenge aims at evaluating plant identification methods and systems at a very large scale, close to the conditions of a real-world biodiversity monitoring scenario. The 2015 evaluation was actually conducted on a set of more than 100K images illustrating 1000 plant species living in West Europe. The main originality of this dataset is that it was built through a large-scale participatory sensing plateform initiated in 2011 and which now involves tens of thousands of contributors. This overview presents more precisely the resources and assessments of the challenge, summarizes the approaches and systems employed by the participating research groups, and provides an analysis of the main outcomes.

## 1 Introduction

Image-based approaches are nowadays considered to be one of the most promising solution to help bridging the botanical taxonomic gap, as discussed in [21] or [16] for instance. We therefore see an increasing interest in this trans-disciplinary challenge in the multimedia community (e.g. in [14], [6], [19], [24], [15], [2]). Beyond the raw identification performances achievable by state-of-the-art computer vision algorithms, the visual search approach offers much more efficient and interactive ways of browsing large floras than standard field guides or online web catalogs. Smartphone applications relying on such image-based identification services are particularly promising for setting-up massive ecological monitoring systems, involving hundreds of thousands of contributors, with different levels of expertise, and at a very low cost.

Noticeable progress in this way was achieved by several projects and apps like LeafSnap[4] [21], PlantNet[5],[6] [16], or Folia[7]. But as promising as these applications are, their performances are however still far from the requirements of a real-world social-based ecological surveillance scenario. Allowing the mass of citizens to produce accurate plant observations requires to equip them with much more accurate identification tools. Measuring and boosting the performances of content-based identification tools is therefore crucial. This was precisely the goal of the ImageCLEF[8] plant identification task organized since 2011 in the context of the worldwide evaluation forum CLEF[9](see [12], [13], [17] and [18] for more details).

---
[4] http://leafsnap.com/
[5] https://play.google.com/store/apps/details?id=org.plantnet&hl=en
[6] http://identify.plantnet-project.org/
[7] http://liris.univ-lyon2.fr/reves/content/en/index.php

Contrary to previous evaluations reported in the literature, the key objective of the PlantCLEF challenge has always been to build a realistic task close to real-world conditions (with many different contributors, cameras, areas, periods of the year, individual plants, etc.). This was initially achieved through a citizen science initiative that began 5 years ago, in the context of the Pl@ntNet project, in order to boost the production of plant images in close collaboration with the Tela Botanica social network. The evaluation dataset was enriched every year with new contributions and progressively diversified with different input feeds (annotation and cleaning of older data, contributions made through Pl@ntNet mobile applications). The plant task of LifeCLEF 2015 was directly in the continuity of this effort. Main novelties compared to the last year were:

- the doubling of the number species, i.e. 1000 species instead of 500
- the possibility to use external training data at the condition that the experiment is entirely re-producible

## 2 Dataset

More precisely, PlantCLEF 2015 dataset is composed of 113,205 pictures belonging to 41,794 observations of 1000 species of trees, herbs and ferns living in Western European regions. This data was collected by 8,960 distinct contributors. Each picture belongs to one and only one of the 7 types of views reported in the meta-data (entire plant, fruit, leaf, flower, stem, branch, leaf scan) and is associated with a single plant observation identifier allowing to link it with the other pictures of the same individual plant (observed the same day by the same person). It is noticeable that most image-based identification methods and evaluation data proposed in the past were so far based on leaf images (e.g. in [21], [3], [6] or in the more recent methods evaluated in [13]). Only few of them were focused on flower's images as in [25] or [1]. Leaves are far from being the only discriminant visual key between species but, due to their shape and size, they have the advantage to be easily observed, captured and described. More diverse parts of the plants however have to be considered for accurate identification, especially because it is not possible for many plant to see their leaves all over the year.

An originality of PlantCLEF dataset is that its "social nature" makes it closer to the conditions of a real-world identification scenario: (i) images of the same species are coming from distinct plants living in distinct areas, (ii) pictures are taken by different users that might not used the same protocol of image acquisition, (iii) pictures are taken at different periods in the year. Each image of the dataset is associated with contextual meta-data (author, date, locality name, plant id) and social data (user ratings on image quality, collaboratively validated taxon name, vernacular name) provided in a structured xml file. The gps geo-localization and device settings are available only for some of the images. Table 1 gives some examples of pictures with decreasing averaged users ratings for the different types of views. Note that the users of the specialized social network creating these ratings (Tela Botanica) are explicitly asked to rate the images according to their plant identification ability and their accordance to the pre-defined acquisition protocol for each view type. This is not an aesthetic or general interest judgement as in most social image sharing sites.

---

[8] http://www.imageclef.org/
[9] http://www.clef-initiative.eu/

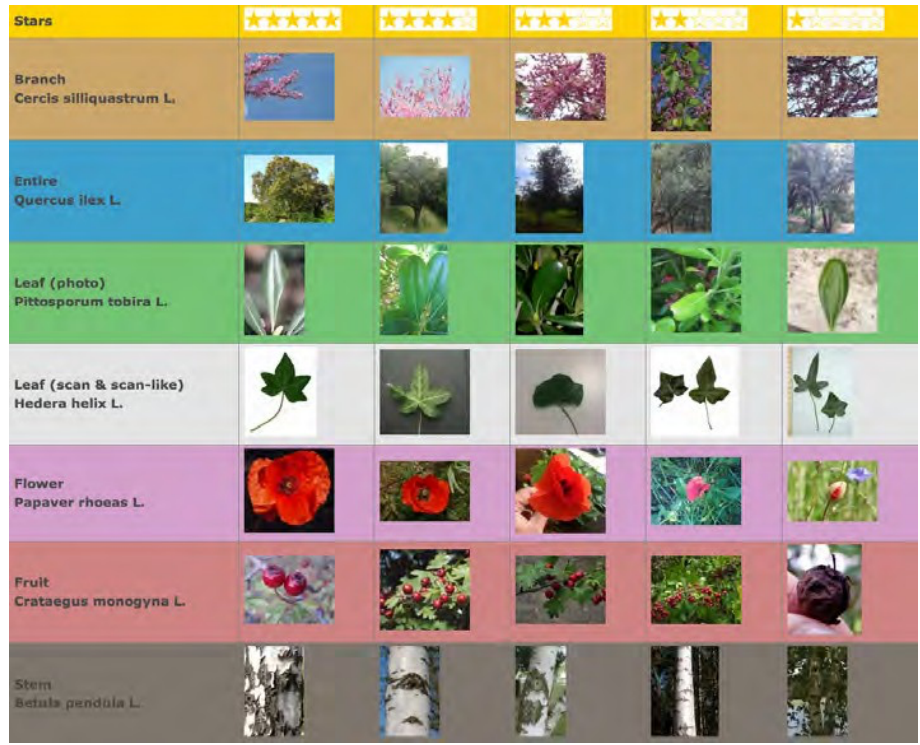

**Fig. 1.** Examples of PlantCLEF pictures with decreasing averaged users ratings for the different types of views

To sum up each image is associated with the followings meta-data:

- **ObservationId**: the plant observation ID from which several pictures can be associated
- **FileName**

- **MediaId**: id of the image
- View **Content**: Branch or Entire or Flower or Fruit or Leaf or LeafScan or Stem
- **ClassId**: the class number ID that must be used as ground-truth. It is a numerical taxonomical number used by Tela Botanica
- **Species** the species names (containing 3 parts: the Genus name, the Species name, the author(s) who discovered or revised the name of the species)
- **Genus**: the name of the Genus, one level above the Species in the taxonomical hierarchy used by Tela Botanica
- **Family**: the name of the Family, two levels above the Species in the taxonomical hierarchy used by Tela Botanica
- **Date**: (if available) the date when the plant was observed,
- **Vote**: the (round up) average of the user ratings of image quality
- **Location**: (if available) locality name, most of the time a town
- **Latitude & Longitude**: (if available) the GPS coordinates of the observation in the EXIF metadata, or, if no GPS information were found in the EXIF, the GPS coordinates of the locality where the plant was observed (only for the towns of metropolitan France)
- **Author**: name of the author of the picture,
- **YearInCLEF**: ImageCLEF2011, ImageCLEF2012, ImageCLEF2013, PlantCLEF2014, PlantCLEF2015 specifying when the image was integrated in the challenge
- **IndividualPlantId2014**: the plant observation ID used last year during the LifeCLEF2014 plant task,
- **ImageID2014**: the image id.jpg used in 2014.

## 3 Task Description

The challenge was evaluated as a plant species retrieval task based on multi-image plant observation queries. The goal was to retrieve the correct plant species among the top results of a ranked list of species returned by the evaluated system. Contrary to previous plant identification benchmarks, queries were not defined as single images but as *plant observations*, meaning a set of one to several images depicting the same individual plant, observed by the same person, the same day, with the same device. Each image of a query observation is associated with a single view type (entire plant, branch, leaf, fruit, flower, stem or leaf scan) and with contextual meta-data (data, location, author).

The whole PlantCLEF dataset was split in two parts, one for training (and/or indexing) and one for testing. All observations with pictures used in the previous plant identification tasks were directly integrated in the training dataset. Then for the new observations and pictures, in order to guarantee that most of the time each species contained more images in the training dataset than in the test dataset, we used a constrained random rule for putting with priority observations with more distinct organs and views in the training dataset. The test set was built by choosing 1/2 of the observations of each species with this constrained random rule, whereas the remaining observations were kept in the reference training set. Thus, 1/3 of the pictures are in the test dataset (see Table 1 for more detailed stats). The xml files containing the meta-data of the *query* images were purged so as to erase the taxon names (the ground truth) and the image quality ratings (that would not be available at query stage in a real-world application). Meta-data of the observations in the training set are kept unaltered.

Table 1: Detailed numbers of images of the LifeCLEF 2015 Plant Task dataset

|       | Total   | Branch | Entire | Flower | Fruit  | Leaf   | LeafScan | Stem   |
|-------|---------|--------|--------|--------|--------|--------|----------|--------|
| Train | 91,759  | 8,130  | 16,235 | 28,225 | 7,720  | 13,367 | 5,476    | 12,605 |
| Test  | 21,446  | 2,088  | 2,983  | 6,113  | 8,327  | 1,423  | 696      | 935    |
| All   | 113,205 | 10,218 | 19,218 | 34,438 | 16,047 | 14,790 | 6,172    | 13,540 |

As a novelty this year, participants to the challenge were allowed to use external training data at the condition that (i) the experiment is entirely reproducible, i.e. that the used external resource is clearly referenced and accessible to any other research group in the world, (ii) participants submit at least one run without external training data so that we can study the contribution of such resources, (iii) the additional resource does not contain any of the test observations. It was in particular strictly forbidden to crawl training data from the following domain names:
http://ds.plantnet-project.org/
http://www.tela-botanica.org
http://identify.plantnet-project.org
http://publish.plantnet-project.org/
http://www.gbif.org/

In practice, each candidate system was evaluated through the submission of a *run*, i.e. a file containing a set of ranked lists of species (each list corresponding to one query observation and being sorted according to the confidence score of the system in the suggested species). Each participating group was allowed to submit up to 4 runs built from different methods. The metric used to evaluate the submitted runs is an extension of the mean reciprocal rank [29] classically used in information retrieval. The difference is that it is based on a two-stage averaging rather than a flat averaging such as:

$$S = \frac{1}{U} \sum_{u=1}^{U} \frac{1}{P_u} \sum_{p=1}^{P_u} \frac{1}{r_{u,p}} \qquad (1)$$

where U is the number of users (within the test set), $P_u$ the number of individual plants observed by the u-th user (within the test set), $r_{u,p}$ is the rank of the correct species within the ranked list of species returned by the evaluated system (for the p-th observation of the u-th user). Note that if the correct species does not appear in the returned list, its rank $r_{u,p}$ is considered as infinite. Overall, the proposed metric allows compensating the long-tail distribution effects occurring in social data. In most social networks, few people actually produce huge quantities of data whereas a vast majority of users (the long tail) produce much less data. If, for instance, only one person did collect an important percentage of the images, the classical mean reciprocal rank over a random set of queries would be strongly influenced by the images of that user to the detriment of the users who only contributed with few pictures. This is a problem for several reasons: (i) the persons who produce the more data are usually the most expert ones but not the most representative of the potential users of the automatic identification tools. (ii) The large number of the images they produce makes the classification of their observations easier because they tend to follow the same protocol for all their observations (same device, same position of the plant in the images, etc.) (iii) The images they produce are also usually of better quality so that their classification is even easier.

A secondary metric was used to evaluate complementary (but not mandatory) runs providing species prediction at the image level (and not at the observation level). The evaluation metric in that is expressed as:

$$S = \frac{1}{U} \sum_{u=1}^{U} \frac{1}{P_u} \sum_{p=1}^{P_u} \frac{1}{N_{u,p}} \sum_{n=1}^{N_{u,p}} \frac{1}{r_{u,p,n}} \quad (2)$$

where $U$ is the number of users, $P_u$ the number of individual plants observed by the $u$-th user, $N_{u,p}$ the number of pictures of the $p$-th plant observation of the $u$-th user, $r_{u,p,n}$ is the rank of the correct species within the ranked list of images returned by the evaluated system.

## 4 Participants and methods

123 research groups worldwide registered to LifeCLEF plant challenge 2015 in order to download the dataset. Among this large raw audience, 7 research groups succeeded in submitting runs on time and 6 of them submitted a technical report describing in details their system. Participants were mainly academics, specialized in computer vision, machine learning and multimedia information retrieval. We list below the participants and give a brief overview of the techniques used in their runs. We remind here that LifeCLEF benchmark is a system-oriented evaluation and not a deep or fine evaluation of the underlying algorithms. Readers interested by the scientific and technical details of any of these methods should refer to the LifeCLEF 2015 working notes of each participant (referenced below):

**EcoUAN (1 run) [27], Colombia**. This participant used a deep learning ap- proach based on a Convolutional Neural Network (CNN). They used the CNN architecture introduced in [20] and that was pre-trained using the popular Im- ageNet image collection [10]. A tuning process was conducted to train the last layer using PlantCLEF training set. The classification at the observation level was done using a sum pooling mechanism based on the individual images classification.

**INRIA-ZENITH (3 runs) [7], France.** This research group experimented two popular families of classification techniques, i.e. convolutional neural networks (CNN) on one side and fisher vectors-based discriminant models on the other side. More precisely, the run entitled INRIA ZENITH Run 1 was based on the GoogLeNet CNN as described in [28] (pre-trained on the popular ImageNet dataset). A single network was trained for all types of view and the fusion of the images of a given observation was performed through a Max pooling. The FV representation used in INRIA ZENITH Run 2 was built from a Gaussian Mixture Model (GMM) of 128 visual words computed on top of different hand-crafted visual features that were previsouly reduced thanks to a Principal Component Analysis (PCA). The classifier trained on top of the FV representations was a logistic regression, which was preferred over a Support Vectors Machine because it directly outputs probabilities which facilitate fusion purposes. INRIA-ZENITH Run 3 was based on a fusion of Run 1 and Run2 using a Bayesian inference framework making use of the confusion matrix of each classifier trained by cross-validation.

**MICA (3 runs) [22], VietNam.** This participant used different hand-crafted visual features for the different view types and trained support vector machines for the classification. The hand-crafted visual features mainly differ in the way the main region of interest is selected before extracting the features:

- For leaf scans, fruit and flower images: automatic selection of a region of interest by using salient features and mean-shift algorithms.
- For leaf images: Segmenting the leaf region by using a watershed algorithm with manual inner/outer markers
- For stem images: Select stem regions by applying a Hanning filter with a pre-determined window size.

The feature extraction step in itself is based on kernel descriptors, namely a gradient kernel for the leafscan, fruit, flower, leaf, entire and branch view type, and a LBP kernel for the stem view type. The late fusion of the SVM classifiers of each view type is based on the sum of the inverse rank position in each ranked list of species. The second run, (run 2) differs from the first one in that it uses complementary HSV histogram features for the flower and entire view types. The third run (Run 3) differs from Run 2 in the fact that it uses an alternative fusion strategy based on a weighted probability combination.

**QUT RV (3 runs), [11], Australia.** This group mainly based his experiment on the use of the GoogLeNet convolutional neural network [28] pre-trained on ImageNet dataset. The 3 runs only differ on the strategy used to fuse the classification results of each image of a query observation (sum pooling in Run 1, softmax in Run 2, normalization & softmax in Run3).

**Sabanki-Okan (3 runs) [23], Turkey.** This group focused its experiment on the evaluation of PCANet [8], a very simple yet efficient deep learning network for image classification which comprises only the very basic data processing components: cascaded principal component analysis (PCA), binary hashing, and block-wise histograms. The original method of [8] was only modified to handle unaligned images. In Run 1, the PCANet is used alone, without using any additional metadata. In Run 2, the date field of the metadata was used to post-process the results of the PCANet. Finally, Run 3 was a trial to combine more classical hand-crafted features for some of the organs (actually SIFT-based VLAD features for Fruit/Leaf/Stem/Branch) with the PCANet approach for the Flower and Entire categories (no meta data used).

**SNUMED (4 runs), [9], Korea.** As the QUT RV and the INRIA ZENITH research groups, the SNUMED group mainly based his experiment on the use of the GoogLeNet convolutional neural network [28] pre-trained on ImageNet dataset. In SNUMED INFO Run 1 and SNUMED INFO Run 2 they fine-tuned a single network across all the whole PlantCLEF 2015 dataset. In SNUMED INFO Run 3 and SNUMED INFO Run 4, they used a different training strategy consisting in randomly partitioning the PlantCLEF training set into five-fold so as to obtain 5 complementary CNN classifier whose combination is supposed to be more stable. The scores at the observation level were obtained by combining the image classification results with the Borda-fuse method.

**UAIC (1 run), [], Romania.** This participant used a content-based image search engine (Lucene Image Retrieval Library []) to retrieve the most similar images of each query image and then apply a two-stage instance-based classifier returning the top-10 most populated species for each image and then the top-10 most populated across all the images of a query observation.

## 5 Results

### 5.1 Main task

The following graphic 2 and table 2 show the scores obtained on the main task (i.e. at the observation level). It is noticeable that the top-9 runs which perform the best were based on the GoogLeNet [28] convolutional neural network which clearly confirms the supremacy of deep learning approaches over hand-crafted features as well as the benefit of training deeper architecture thanks to the improved utilization of the computing resources inside the network. The score's deviations between these 9 runs are however still interesting (actually 10 points of mAP between the worst and the best one). A first source of improvement was the fusion strategy allowing to combine the classification results at the image level into classification scores at the observation level. In this regard, the best performing algorithm was a SoftMax function [4] as shown by the performance of QUT RV Run 2 compared to INRIA ZENITH run1 based on max pooling, or SNUMED INFO run1 based on a Borda count, or QUT RV run1 based on a sum pooling. The other source of improvement, which allowed the SNUMED group to get the best results, was to use a bootstrap aggregating (bagging) strategy [5] to improve the stability and the accuracy of the GoogLeNet Convolutional Neural Network. In SNUMED INFO Run 3 and SNUMED INFO Run 4, they actually randomly partitioned the PlantCLEF training set into five-fold so as to train 5 complementary CNN classifiers. Bagging is a well known strategy for reducing variance and avoiding overfitting, in particular in the case of decision trees, but it is interesting to see that it is also very effective in the case on deep learning.

Table 2: Results of the LifeCLEF 2015 Plant Identification Task. Column "Key-words" attempts to give the main idea of the method used in each run.

| Run name | Key-words | Score |
|---|---|---|
| SNUMED INFO run4 | 5-fold GoogLeNet Borda+ | 0,667 |
| SNUMED INFO run3 | 5-fold GoogLeNet Borda | 0,663 |
| QUT RV run2 | GoogLeNet SoftMax | 0,633 |
| QUT RV run3 | GoogLeNet Norm & SoftMax | 0,624 |
| SNUMED INFO run2 | GoogLeNet Borda+ | 0,611 |
| INRIA ZENITH run1 | GoogLeNet Max Pool. | 0,609 |
| SNUMED INFO run1 | GoogLeNet Borda | 0,604 |
| INRIA ZENITH run3 | Fusion GoogLeNet & Fisher Vectors | 0,592 |
| QUT RV run1 | GoogLeNet Sum Pool. | 0,563 |
| ECOUAN run1 | CNN Sum Pool. | 0,487 |
| INRIA ZENITH run2 | hand-crafted features + Fisher Vectors | 0,300 |
| MICA run2 | Hand-crafted feat. + SVM | 0,209 |
| MICA run1 | Hand-crafted feat. + SVM | 0,203 |
| MICA run3 | Hand-crafted feat. + SVM | 0,203 |
| SABANCI run2 | PCANet (not pretrained) | 0,162 |
| SABANCI run1 | PCANet (not pretrained) | 0,160 |
| SABANCI run3 | PCANet (not pretrained) | 0,158 |
| UAIC run1 | CBIR (LIRE) | 0,013 |

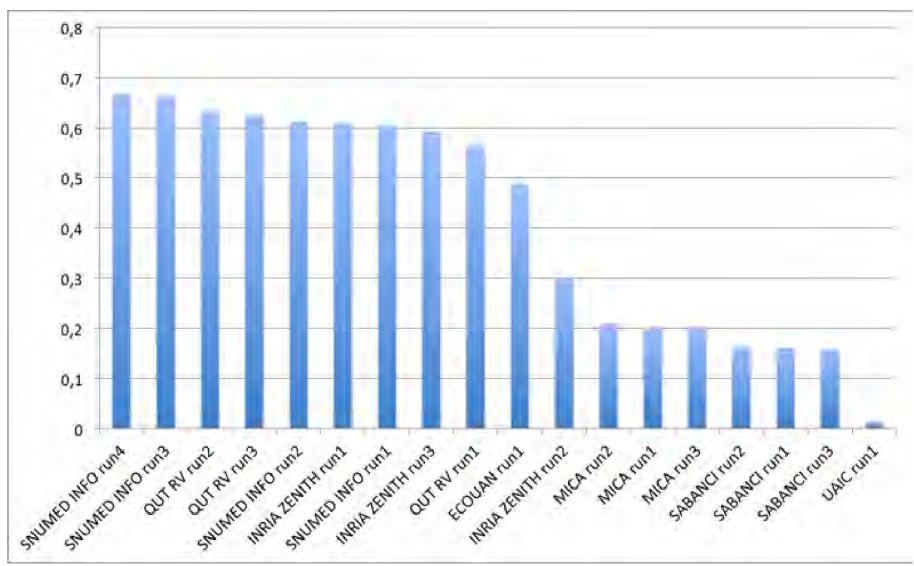

**Fig. 2.** Official results of the LifeCLEF 2014 Plant Identification Task.

The second best approach that did not rely on deep learning (i.e. INRIA ZENITH run 2) was to use the Fisher Vector model [26] on top of a variety of hand-crafted visual features and then to train a multi-class supervised linear classifier through logistic regression. It is here important to note that this method does not make use of any additional training data other than the one provided in the bench- mark (contrary to the CNN's that were all previously trained on the large-scale ImageNet dataset). Within the 2014 PlantCLEF challenge [18], in which using external training data was not allowed, the Fisher Vector approach was per- forming the best, even compared to CNN's. But still, the huge performance gap confirms that learning visual features with deep learning is much more effective than sticking on hand-crafted visual features. Interestingly, the third run of the INRIA ZENITH team was based on a fusion of the fisher vector run and the GoogLeNet one which allows assessing in which measure the two approaches are complementary or not. The results show that the performance of the fused run was not better than the GoogLeNet alone. This indicates that the hand-crafted visual features encoded in the fisher vectors did not bring sufficient additional information to be captured by the fusion model (based on Bayesian inference). A last interesting outcome that can be derived from the raw results of the task is the relative low performance achieved by the runs of the SABANCI research group which were actually based on the recent PCANet method [8]. PCANet is a very simple deep learning network which comprises only basic data pro- cessing components, i.e. cascaded principal component analysis (PCA), binary hashing, and block-wise histograms. The learned visual features are claimed by the authors to be on par with the state of the art features, either prefixed, highly hand-crafted or carefully learned (by DNNs). The results of our challenge do not confirm this assertion. All the runs of SABANCI did notably have lower per- formances than the hand-crafted visual features used by MICA runs or INRIA ZENITH Run 2, and much lower performances than the features learned by all other deep learning methods. This conclusion should however be mitigated by the fact that the PCANet of SABANCI was only trained on PlantCLEF data and on a large-scale external data such as ImageNet. Complementary experiments in this way should therefore be conducted to really conclude on the competitiveness of this simple deep learning technique.

## 5.2 Complementary results on images

The following graphic 3 presents the scores obtained by the additional image-level runs provided by the participants. In order to evaluate the benefit of the combination of the test images from the same observation, the graphic compares the pairs of run files on images and on observations produced with the same method.

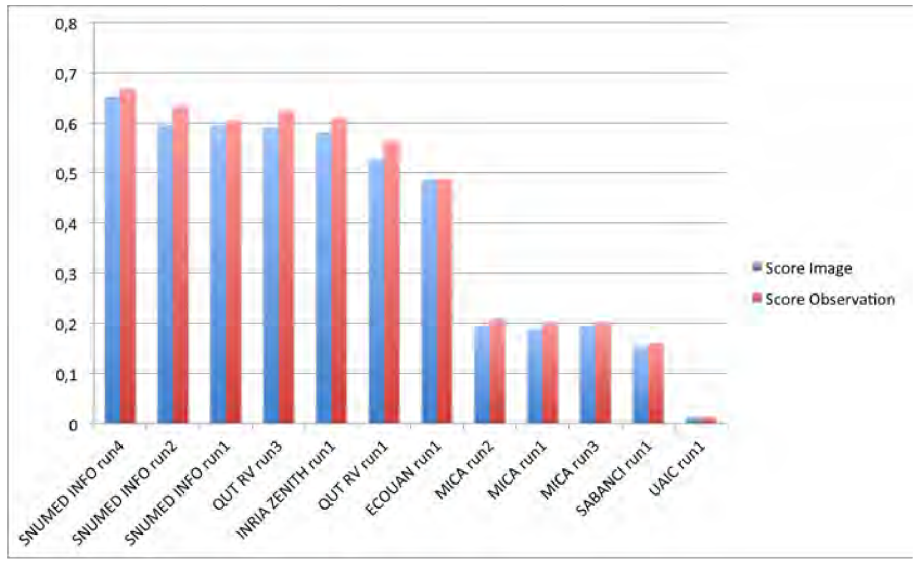

**Fig. 3.** Comparison of the methods: before and after combining the prediction for each image from a same plant observation.

Basically, for each method, we can observe an improvement by combining the different views of the same plant observation. This has to be related to the fact that observing different plant organs is the current practice of botanists, who most of the time can't identify a species with only one picture on only one organ. However, we can say that the improvement are not so much high: we guess that there is a room of improvement here, basically with more images and may be with new methods of fusions dealing with this specific problem of multi-image and multi-organ problem.

## 5.3 Complementary results on images detailed by organs

The following graphic 4 below show the detailed scores obtained for each type of organs. Remember that we use a specific metric weighted by authors and plants, and not by sub-categories, explaining why the score on images not detailed is not the mean of the 7 scores of these sub-categories.

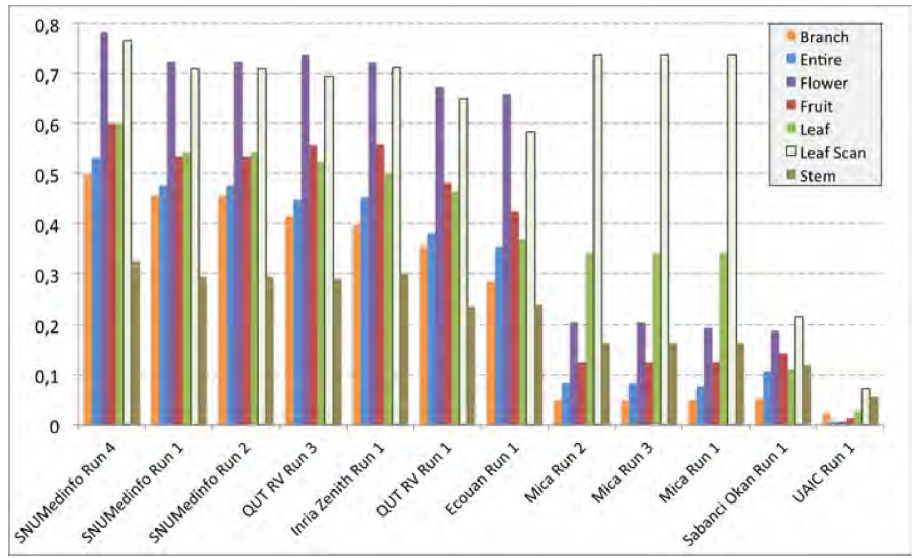

**Fig. 4.** Results detailed for each type of image category.

Like during the previous Plant Identification Task, this detailed analysis shows that LeafScan and the Flower views are far away the most effective for identifying plant species, followed by the Fruit view, the Leaf view, the Entire view and the Branch view. On the other side, the stem view (or bark view when speaking about trees) is the less informative one particularly when noticing that the number of species represented in that view, and thus the confusion risk, was lower than for the other organs. Interestingly, the hand-crafted visual features of the MICA group perform very well on the Leaf Scan category, with an identification score better than most of the runs based on the GoogLeNet CNN. This shows the relevance of the leaf normalization strategy they used as well as the effectiveness of the gradient kernel for this type of view. For professional use cases in which taking the time to scan the leaf might not be an issue, this method is a serious alternative to the use of the CNN which requires much more resources and training data.

## 6  Conclusion

This paper presented the overview and the results of the LifeCLEF 2015 plant identification challenge following the four previous ones conducted within CLEF evaluation forum. The main novelty compared to the previous year was the possibility to use external training data in addition to the specific training set provided within the testbed. The first objective of this novelty was clearly to encourage the deployment of transfer learning methods, and in particular of deep convolutional neural networks in order to evaluate their ability to identify plant species at a large-scale. In this regard, the results show that such transfer learning approaches clearly outperform previous approaches based on hand-crafted visual features, aggregation models and linear classifiers. The results are as impressive as a $0,784$ identification score on the flower category. Now, the second objective of opening the training data was to encourage the integration of new plant data (and not only of the popular generlist dataset ImageNet), particularly for populating the long tail of the less populated species which is an important challenge in terms of biodiversity. Unfortunately, none of the participants addressed this issue. More generally, we believe that collecting and building appropriate training data is becoming one of the most central problem for solving definitely the taxonomic gap problem.